\documentclass{article}
\usepackage{lipsum}
\usepackage{graphicx}
\usepackage{multirow}
\usepackage{makecell}
\usepackage{xspace}
\usepackage{rotating}
\usepackage{arxiv}

\usepackage{microtype}
\usepackage{graphicx}
\usepackage{subfigure}
\usepackage{booktabs}
\usepackage{tikz}
\usepackage{arxiv}
\usepackage{siunitx}

\usepackage{algorithm}
\usepackage{algpseudocodex}
\RequirePackage{fancyhdr}
\RequirePackage{xcolor}
\RequirePackage{natbib}
\RequirePackage{eso-pic}
\RequirePackage{forloop}
\RequirePackage{url}

\usepackage{amsmath}
\usepackage{amssymb}
\usepackage{mathtools}
\usepackage{amsthm}
\usepackage{mathrsfs}

\newcommand{\hpa}{\unit{\hecto\pascal}\xspace}
\newcommand{\hr}{\unit{\hour}\xspace}
\newcommand{\dd}{\unit{\day}\xspace}

\renewcommand{\shorttitle}{Fine-tuning GraphCast with GDPS}

\title{Efficient fine-tuning of 37-level GraphCast with the Canadian global deterministic analysis}
\author{Christopher Subich\thanks{Corresponding author, Environment \& Climate Change Canada, Dorval Québec} \\
\texttt(christopher.subich@ec.gc.ca)}

\begin{document}
\maketitle

\begin{abstract}
    This work describes a process for efficiently fine-tuning the GraphCast data-driven numerical weather prediction model to be consistent with another analysis system, here the Global Deterministic Prediction System (GDPS) of Environment and Climate Change Canada (ECCC).  The GDPS system was significantly upgraded in 2019, so there is only a limited period of operational data to use for model training.  This work considers the effect of using two years of training data (July 2019--December 2021) and a restricted computational budget to tune the 37-level, quarter-degree version of GraphCast with an empirically-determined vertical weighting of model error.  The GDPS-tuned model significantly outperforms both the operational (traditional) forecast and the unmodified GraphCast model when initialized with the GDPS analysis, showing significant forecast skill in the troposphere over 1 to 10-day lead times.  This fine-tuning is accomplished through an abbreviation of the original training curriculum, relying on a shorter single-step forecast stage to complete most of the adaptation, followed by a consolidation of the forecast-lengthening stages into separate 12\hr, 1\dd, 2\dd, and 3\dd stages.  In addition, a ``control'' run trained with ERA5 data shows that fine-tuning on recent data improves forecast skill on a going-forward basis.
\end{abstract}

\section{Introduction}

The development of machine learning methods for global atmospheric prediction with resolution comparable to operational global models has begun a revolution in medium-range weather prediction.  GraphCast \citep{graphcast}, FourCastNet \citep{fourcastnetv1,fourcastnet}, and Pangu-Weather \citep{pangu} all demonstrated forecast skill superior to that of the European Centre for Medium-Range Weather Forecasts's (ECMWF's) high resolution model (HRES) at lead times up to ten days.

Thanks to the comprehensive use of accelerator cards (generally graphics processing units (GPUs)), these data-driven models provide forecasts in a fraction of the computational time required for traditional models at comparable resolutions.

Some commenters \citep{future_ai} expect this revolution to continue, where one day data-driven forecasting will entirely supplant conventional numerical weather prediction.  However, today is not that day.  National and international weather centres are deploying these models and their own \citep{aifs} alongside their traditional systems, providing experimental or auxiliary outputs.  The output of data-driven models has also shown utility as guidance for longer-range traditional forecasts \citep{nudging}, combining the improved large-scale accuracy of data-driven models with the higher-resolution and more comprehensive output of the traditional forecasting system.

While data-driven models supplement rather than replace traditional forecasting, weather centres will need to have models that are adapted to their local operational configurations.  Most published global models have been trained on the ECMWF Reanalysis v5 (ERA5) dataset \citep{era5}, notable for both its high quality and its extended period of availability.  Unfortunately, data-driven models tend to suffer performance degradation when their input data is out of distribution with respect to the training data.  Although all operational forecasting systems attempt to model the same Earth, deployed systems such as Environment and Climate Change Canada's (ECCC's) Global Deterministic Prediction System (GDPS) can have systematic differences with respect to ERA5 and its underlying Integrated Forecasting System (IFS) model.

Therefore, national centers seeking to deploy data-driven forecasting systems are faced with a choice:
\begin{enumerate}
    \item They can rely on the ERA5 reanalysis for short-term forecasting, despite it being available only with several days' delay,
    \item They can apply a data-driven forecast model to their operational analysis and accept any resulting performance degradation, or
    \item They can use an in-house version of a forecast model, either by developing such a system ``from scratch'' or by fine-tuning an already-existing model.
\end{enumerate}

This work elaborates on the third option, discussing the steps required to fine-tune GraphCast for the operational Canadian GDPS analysis using a relatively short training period taken from the operational system, without needing a long reanalysis data series.  The primary endpoint of the study is to develop a set of model weights that will predict a future GDPS analysis when provided with compatible initial conditions, over lead times from six hours to ten days when run autoregressively.  

\subsection{Prior work}

The idea of fine-tuning a data-driven model for an operational forecast system is not new, and some efforts have been made already.  Google DeepMind itself has published a 13-level version of the GraphCast model (\citet{graphcast_github}, checkpoint \texttt{graphcast\_operational}) that was fine-tuned on ECMWF HRES data for the 2016--2021 period.  The National Oceanic and Atmospheric Administration has also taken these weights and fine-tuned them with initial conditions from its Global Deterministic Analysis System for initial conditions between March 2021 and September 2022 \citep{noaa_graphcast}.  However, technical details for reproduction of these efforts have not been fully published.

To the author's knowledge, this work the first detailed discussion of fine-tuning for the 37-level, quarter-degree (full) version of GraphCast.

\subsection{Organization}

Section~\ref{sec:ics} reviews the main features of GraphCast and the GDPS data used for fine-tuning and the computational environment used.  Section~\ref{sec:method} discusses the design choices made for the fine-tuning process, section~\ref{sec:results} shows the accuracy over the training process and validation results on a hold-out period, and section~\ref{sec:conclusion} concludes with further discussion about the general applicability of this procedure for other centres or datasets.

\section{The model and datasets}\label{sec:ics}

\subsection{GraphCast}

GraphCast \citep{graphcast} is based on a graph neural network architecture, with separate encoding, processing, and decoding stages.  The encoding and decoding stages each act on a quarter-degree latitude/longitude grid, but the processing stage propagates data on a multi-resolution, quasi-uniform icosahedral mesh with approximately one-degree vertex spacing.  This work is based on the 37-level version of GraphCast\footnote{\citet{graphcast_github} also provides weights for a one-degree, reduced resolution version of GraphCast and a 13-level version that has been fine-tuned on operational HRES data, but the training details are not discussed in \citet{graphcast}.}, which features 16 processor stages and about 37 million trainable parameters.  The variables predicted by GraphCast and required for its successful execution are described in table~\ref{tab:gc_vars}.

GraphCast has relatively few trainable parameters compared to other global models.  Pangu-Weather has about 256 million trainable parameters, while FourCastNet has about 433 million, both to represent 13 atmospheric model levels plus surface variables.  Meanwhile, the 37-level, quarter degree atmospheric output consists of about 235 million values.  With nearly an order of magnitude fewer trainable parameters, GraphCast does not have enough degrees of freedom to memorize even a single training example at full precision.  Thus, GraphCast should be relatively resistant to over-fitting during training or fine-tuning, even over relatively small datasets.

\begin{table}
    \begin{center}
    \begin{tabular}{cc}
        \multicolumn{2}{c}{\textbf{Input and output variables}}  \\
        Surface & Three-dimensional  \\ \hline
        \makecell{2-meter temperature (t2m) \\ 6h-accumulated precipitation (tp) \\ 10-meter zonal wind (u10m) \\ 10-meter meridional wind (v10m)\\ Mean sea level pressure (msl)} &
        \makecell{Geopotential (z) \\ Specific humidity (q) \\ Temperature (t) \\ Zonal wind (u) \\ Meridional wind (v) \\ Vertical wind speed (w)}  \\ \\ 
        \multicolumn{2}{c}{\textbf{Time-dependent input-only variables}} \\ 
        Variable & Type \\ \hline
        \makecell{1h-accumulated solar radiation} & Two-dimensional \\
        \makecell{Cosine of day fraction \\ Sine of day fraction} & One-dimensional (longitude) \\
        \makecell{Cosine of year fraction \\ Sine of year fraction} & Scalar \\ \\
        \multicolumn{2}{c}{\textbf{Time-independent input-only variables}} \\
        Variable & Type \\ \hline
        \makecell{Land-sea mask \\ Surface topography (geopotential)} & Two-dimensional \\ \\ 
        \multicolumn{2}{c}{\textbf{Implicit input features}} \\
        Grid/mesh nodes & Mesh edges \\ \hline
        \makecell{Sine of latitude \\ Cosine of longitude \\ Sine of longitude} &
        \makecell{Edge distance \\ Directed displacement}
    \end{tabular}
    \end{center}
    \caption{Input and output variables for GraphCast, from \citet{graphcast}}\label{tab:gc_vars}
\end{table}

\subsubsection*{Computational considerations}

GraphCast was originally trained on Google's computational cloud, using its proprietary Tensor Processing Unit (TPU) accelerators.  Environment \& Climate Change Canada, however, has internal access to several high-performance computing nodes, each containing four NVIDIA A100 GPUs with 40 gigabytes of RAM apiece.  Each node contains two Intel Xeon Gold 6342 CPUs (24 cores) and 500 gigabytes of system RAM, of which about 400 gigabytes are usable for user jobs.

These nodes are more than capable of making predictions with GraphCast's published code when given existing model weights, but training GraphCast at a quarter-degree resolution requires code modifications.  
Although the model's inference code was open-sourced by \citet{graphcast}, the training code was not released, and a straightforward implementation of a training loop requires more than 40 gigabytes of GPU memory to compute gradients for a single, 6\hr forecast step, making training effectively impossible on ECCC's systems.

Lam et al.~kept their training code proprietary because it relied on internal infrastructure and specific gradient checkpointing and memory-offloading strategies\footnote{\texttt{https://github.com/google-deepmind/graphcast/issues/55\#issuecomment-1910222399}, retrieved 18 July 2024}. Both of these were effectively replicated on GPU for this study.  Gradient checkpointing calls were conservatively added to each nonlinear activation step inside the graph neural network, and data was implicitly offloaded to system RAM as-needed through unified memory, which allows processes to transparently mix CPU and GPU memory for computations.  

The net result was that backpropagation over a single forecast step was possible within the confines of primary GPU memory, and backpropagation up to eight steps (two days) was possible with an acceptable performance penalty by enabling unified memory.  Beyond that, system memory limits effectively prevented longer forecast horizons, but backpropagation on twelve-step (three day) forecasts was possible through split-horizon training (see section~\ref{subsec:split}).  More host memory would have made this concession unnecessary, and more GPU memory would have likely significantly improved training speed.

For its principal training, GraphCast was trained with batch-level data parallelism, where each member of a training batch had its gradients computed by a separate accelerator.  The present work shares that approach, and each GPU was given its own computational Python thread.  In testing, this allowed for maximum overlap of data loading, processing, and GPU computation, and training was moderately faster than an alternative implementation that used data-parallel sharding in the JAX library \citep{jax}.  Each GPU worker thread computed gradients over separate samples, and these gradients were aggregated after each batch.

\subsubsection*{Computational budget}

\citet{graphcast} reports that the 37-level, quarter-degree version of GraphCast took about four weeks to train on 32 TPU accelerators, corresponding to about 2.5 accelerator-years of computational time.  This is a small fraction of the computational time used to train very large language models.  However, this is still a significant investment that might effectively prohibit following updates to an operational analysis system with corresponding updates to the data-driven emulator.  To consider the effect of computational constraints, this study sets a self-imposed limit of about six GPU-weeks, which prohibits simple replication of the published GraphCast training schedule for fine-tuning.

\subsection{Training and test data}

The main dataset used in this study is the ``late'' operational analysis produced by the Global Deterministic Prediction System.  On July 3, 2019, version seven of this system became operational \citep{gdps7,gdps_physics}, which upgraded the atmospheric component of the forecast to a nominal fifteen kilometer horizontal resolution, based on a quasi-uniform Yin-Yang grid \citep{yinyang} consisting of two overlapping panels.  Although the system produces several atmospheric analyses for operational forecasts with various real-time deadlines, the ``late'' analysis has a data cutoff time of +7 hours after its nominal initialization time (e.g. the 0h UTC forecast incorporates data up to 7h UTC).  

This analysis field is produced four times daily, at 0h, 6h, 12h, and 18h UTC.  With a seven-hour data cutoff it is obviously not available for real-time forecasting, but it provides the most accurate representation of the atmospheric state available operationally.  The extended data cutoff is comparable to the 12h analysis window used for ERA5, which has a similar problem of ``seeing into the future'' for some of its hourly products.  Nonetheless, this dataset was ideal for training because of its consistency and quality.  The operational short-cutoff analysis is only available at 0h and 12h UTC, so using this for training would have still required another dataset for 6h and 18h UTC conditions.

This version of GDPS was operational until December 1, 2021, whereupon a subsequent upgrade \citep{gdps8} included substantial updates to the analysis component.  Consequently, July 2019 -- December 2021 provides a natural training period.  Calendar year 2022 was held out as a validation dataset during development of the procedure described in this article, and calendar year 2023 was held out as a test set for the results in section \ref{sec:results}.  In contrast, the unmodified version of GraphCast was trained on ERA5 data from the 1979--2017 period.


The higher-resolution analysis data was interpolated to the quarter-degree latitude/longitude grid via conservative remapping with the xESMF Python library \citep{xesmf}.  Both of the grid panels were interpolated to the globe separately, with the simple average being taken in the regions where the two panels overlap.  

Vertical velocity and accumulated precipitation are not present in the GDPS analysis files, so these values were taken from the preceding ``early'' forecast\footnote{This forecast is initialized every six hours beginning at midnight UTC, and it provides boundary conditions for the regional forecasting systems.  The primary operational forecast is only initialized at midnight and noon UTC, so using the early forecast gave a more consistent training dataset.} at six hours' lead time.  As a concrete example, in the training dataset the vertical velocity and accumulated precipitation fields valid on July 1, 2021, at 12:00 UTC were those given by the GDPS early forecast initialized on July 1, 2021, at 6:00 UTC. 

Consequently, vertical velocity and accumulated precipitation should be interpreted cautiously.  Fortunately, an extension of the sensitivity analysis in section~\ref{subsec:errweights} suggests that GraphCast does not heavily rely on these variables for future predictions, and instead they are mostly controlled by the other variables.

\subsubsection*{A control case}

The limited operational period of the GDPS analysis causes this study to conflate two factors in the fine-tuning: the use of GDPS rather than ERA5 for training data and the use of a restricted time period for training data.  To help separate these factors, this study also includes a ``control case'' where the same fine-tuning procedure is applied to training data from the ERA5 dataset.

This control case uses the same date ranges, with the primary training set consisting of July 2019 through December 2021, a validation dataset of calendar year 2022 used for validation, and a held-out test set of calendar year 2023.  ERA5 data through 2022 was downloaded from the WeatherBench~2 dataset \citep{weatherbench2}, and data for calendar year 2023 was downloaded from ECMWF's Climate Data Store.

The restricted but modern period for fine-tuning may affect the accuracy of GraphCast forecasts in three partially offsetting ways:
\begin{enumerate}
    \item Using a limited fine-tuning period might cause the model to forget important modes of variability that only exist in the historic dataset.  Models such as \citet{climax} and \citet{aurora} claim improvements from the incorporation of extra training data diversity such as climate simulations, which is essentially the opposite of fine-tuning over a restricted dataset.
    \item Using data close to the evaluation time might cause accuracy improvements from implicit learning of unrepresented variables.  For example, GraphCast does not receive or predict ocean temperatures, land surface usage (including albedo), or the sea ice mask, and a model trained for a short period might be able to better-infer the influence of these variables through learned short-term correlations that do not hold over decades-long time scales.  This may also allow for a better representation of processes affected by climate change, since GraphCast does not receive the forecast year or atmospheric composition as input variables.
    \item Finally, the increasing quality and quantity of observational data over time might have improved the analysis fields, resulting in higher-quality initial conditions that introduce more predictability to the data-driven model.  Earlier portions of the ERA5 dataset are relatively less constrained by observational data, so using the whole period as training data may implicitly ask GraphCast to interpolate between ``what an unconstrained IFS predicts'' and ``what happens in the real atmosphere.''
\end{enumerate}

\citet{graphcast} reports accuracy gains from expanding the training set (initially limited to 2015, but extended to 2017 the published model weights) to include more recent years, but since it expands the whole training set it is unable to differentiate between these factors.  \citet{couairon2024} also found that there was a ``distribution shift'' in ERA5 data after the year 2000, whereby training on more recent data improved their model's predictive ability; the authors attributed this to improvements in the global observation system.

\section{Method}\label{sec:method}

\begin{table}
    \begin{center}
        \begin{tabular}{cccccc}
            Name & Length & Peak learning rate & \# samples & Batch time (\unit{\sec}) & Total time (\hr)\\
            \hline
            GC37     & \multicolumn{4}{c}{Baseline (no training)} \\
                     & (1a) (6\hr)  & $1.25 \cdot 10^{-4}$ & 10240 & 4.54 & 3.23 \\
            Tuned-6h & (1b) (6\hr)  & $3.75 \cdot 10^{-7}$ & 81920 & 5.45 & 31.0 \\ 
                     & 2  (12\hr) & $1.25 \cdot 10^{-6}$ & 20480 & 17.45 & 24.82 \\
            Tuned-1d & 4  (24\hr) & $1.25 \cdot 10^{-6}$ & 10240 & 31.36 & 22.30 \\
            Tuned-2d & 8  (48\hr) & $1.25 \cdot 10^{-7}$ & 10240 & 86.23 & 61.32 \\
            Tuned-3d & 12 (72\hr) & $3.75 \cdot 10^{-7}$ & 10240 & 114.9 & 81.71
        \end{tabular}
    \end{center}
    \caption{Forecast length, peak learning rate, and runtime characteristics for the fine-tuning schedule of this work.  The (1a) and (1b) lines refer to the training stages of sections~\ref{subsec:norms} and~\ref{subsec:errweights} respectively; each stage trains over one step.  The total training time was about 224\hr, or 37.4 GPU-days.}\label{tab:schedule}
\end{table}

\subsection{Overview}

The fine-tuning training of this work followed a multi-stage curriculum summarized in table \ref{tab:schedule}.  To accommodate the full set of differences between the ERA5 and GDPS datasets, training began with single-step forecasts and was extended to multistep forecasts, ultimately reaching the same 72-hour forecast length trained in \citet{graphcast}, skipping certain intermediate steps.  Models whose error properties are shown in the subsequent figures are named in the first column of table \ref{tab:schedule}, and each tuned model is only evaluated against its own dataset.  The baseline model is evaluated with both datasets, and these evaluations will be denoted GC37-GDPS and GC37-ERA5 when noted for clarity.

Training was conducted over mini-batches of size 4, which was a more natural fit for the 4-GPU computational nodes available at training time.  Preliminary testing showed that this choice does not have a large impact on overall performance, provided the leaning rate is scaled appropriately.  Section \ref{subsec:lr} discusses the learning rate in more detail.

\subsection{Optimizer and loss function}

Like GraphCast, the fine-tuning procedure used the AdamW optimizer \citep{adam,adamw}, implemented in the Optax library, which is part of the JAX ecosystem \citep{deepmind2020jax}.  The momentum parameters $\beta_1 = 0.9$ and $\beta_2 = 0.95$ were left unchanged from \citet{graphcast}, as was the weight decay factor of $0.1$.  Gradient clipping was not implemented, as it was not necessary for stability during the fine-tuning cycles.


As with GraphCast's principal training, the training objective was a scalar mean-squared error that combined errors by predicted variable, model level, and forecast lead time.  Reproducing equation~(19) of \citet{graphcast} with small notation changes, the per-forecast error is given by:
\begin{equation}\label{eqn:trainerr}
    \mathrm{MSE} = \underbrace{\sum_{\tau=1}^{N_t} \frac{1}{N_t}}_{\text{Lead time}} 
                   \underbrace{\sum_{i,j} \frac{dA(i,j)}{4\pi}}_{\text{Space}}
                   \underbrace{\sum_{k=0}^{N_k} w(k)}_{\text{Level}} \sum_{\mathrm{var}}
                   \omega_\mathrm{var} 
                   \frac{ (\hat{x}_\mathrm{var}(i,j,k;\tau) - x_\mathrm{var}(i,j,k;\tau))^2}{\sigma^2_{\Delta\mathrm{var}}(k)},
\end{equation}
where $N_t$ is the length of the forecast (in 6-hour increments); $dA(i,j)$ is the angular area of the $(i,j)$th grid cell; $N_k$ is the number of vertical pressure levels (indexed by $k$, with the surface as index zero); $w(k)$ is a per-pressure-level error weight (taking $w(0) = 1$ at the surface); $\omega_{\mathrm{var}}$ is a fixed per-variable error weight; $x_\mathrm{var}(i,j,k;\tau)$ is the model's prediction at grid cell $(i,j)$, vertical level $k$, and lead time $\tau$ for variable $\mathrm{var}$; $\hat{x}$ is the corresponding value from the training analysis field; and $\sigma^2_{\Delta\mathrm{var}}(k)$ is the variance of 6-hour analysis differences over a climatological period, per pressure level and per predicted variable.

This fine-tuning preserves the per-variable weights from GraphCast's principal training: $\omega$ was $1$ for the three-dimensional atmospheric variables and 2-meter temperature, and it was $0.1$ for the 10-meter wind components, mean sea level pressure, and accumulated precipitation.

\subsubsection*{Cosine annealing}

The fine-tuning process consists of several training stages.  \citet{graphcast} describes the principal training of GraphCast as a single, largely continuous affair consisting of a primary single-step stage and a secondary fine-tuning step over (an increasing number of) multiple stages, but this might not be the most appropriate view.  Each stage of training here involves some kind of discontinuous change to the system, whether it be changing the normalization factors (section~\ref{subsec:norms}), error weights (\ref{subsec:errweights}), or forecast length (\ref{subsec:schedule}).  

Therefore, it seems appropriate to give each stage its own learning rate schedule. To that end, this work follows the principal training of GraphCast by using half-cosine learning weight decay to determine per-batch learning weights, with a linear warm-up period over the first 10\% of the stage.  The peak learning rate varied by training stage, and the terminal learning rate was always $3.75\cdot10^{-8}$, the equivalent of the $3\cdot10^{-7}$ learning rate used in \citet{graphcast} linearly scaled to the batch size of 4.

\subsection{Learning rate determination}\label{subsec:lr}


\begin{figure}[tb]
    \begin{center}
        \includegraphics[width=6in]{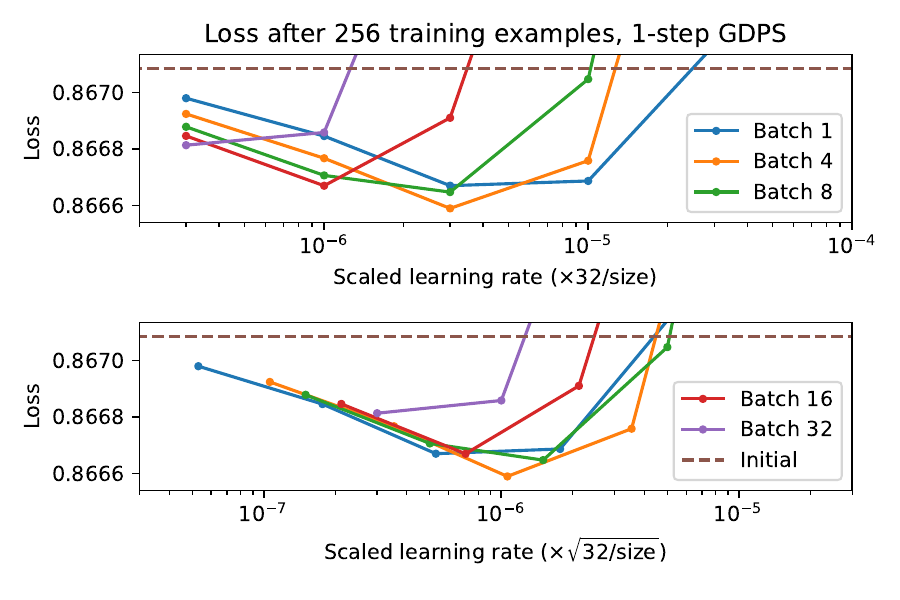}
    \end{center}
    \caption{Validation loss after 256 training examples for the one-step forecasting stage described in section~\ref{subsec:errweights}, for a selection of learning rates and batch sizes.  To show scaling versus learning rate, the top panel normalizes the learning rate by the inverse of the batch size, and the bottom panel normalizes by the inverse of the square root of batch size.  For the batch size of four used in this work, the optimum learning rate was about $3.75\cdot10^{-7}$.}\label{fig:lr_anneal}
\end{figure}

The learning rate for a training stage is obviously important.  A learning rate that is too conservative will waste computational resources, with the model learning less than it might for a given number of training examples.  On the other hand, a learning rate that is too aggressive will result in a model that spends part of the training cycle diverging rather than converging towards its optimum, risking progress made thus far.

Selecting an ideal learning rate is an art, and this work proceeds with a basic search.  The half-cosine learning rate schedule fixes the form and terminal value of the learning rate, so the remaining degree of freedom is the initial (post-warm-up) value that serves as its maximum.  To find this learning rate for every stage but ``1a'' in table \ref{tab:schedule}, the model was trained with several fixed (constant) learning rates for a few training examples (typically 256), then evaluated over 64 forecasts randomly selected from the validation dataset.  The learning rate that provides the smallest validation error at the end of the training was chosen as the peak learning rate for the full stage.

In this work, the learning rate search was applied at each stage with a batch size of 4, and the optimum was used for the various training stages in table \ref{tab:schedule}.  Figure \ref{fig:lr_anneal} shows this process applied to stage 1b from that table for a variety of batch sizes, showing how an optimum learning rate may transfer between different computational configurations.  For low learning rates, the $1/\sqrt{\mathrm{size}}$ scaling rule advanced by \citet{granziol2022} for problems with momentum-based optimizers seems to hold, but the exact location of the best learning rate does not quite follow that curve.  A $1/\mathrm{size}$ scaling might be more appropriate for small batch sizes, but the search was not dense enough to be conclusive. 

Regardless of scaling law, the optimal learning rate for small batch sizes leads to slightly more rapid learning than is found for a batch size of 32, suggesting that there is some partial compensation to be found when training the model with relatively few GPUs.

\subsection{Re-normalizing inputs}\label{subsec:norms}

\begin{figure}[tb]
    \begin{center}
        \includegraphics[width=6in]{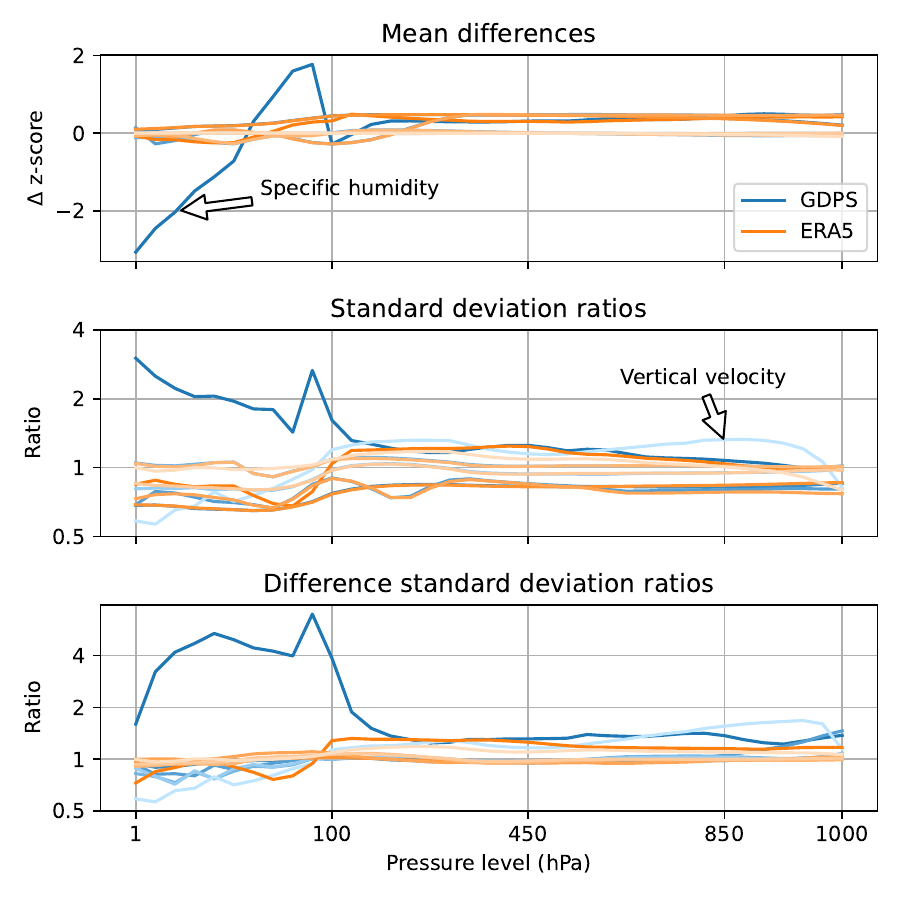}
    \end{center}
    \caption{Differences in the normalization constants when computed over the July 2019 -- July 2021 period compared to the 1979--2015 period for all 3D atmospheric variables, by level.  Mean differences are normalized to z-scores by dividing by the ERA5 1979--2025 standard deviation, and the standard deviations are shown as ratios to the 1979--2015 values.  The limited modern period for fine-tuning causes minor differences in these quantities, but the GDPS dataset (blue lines) shows a large, systematic difference for specific humidity in the stratosphere.} \label{fig:dstd_6h}
\end{figure}

The GraphCast model operates in a normalized sense.  Rather than receive raw fields whose numerical values vary over several orders of magnitude, the initial conditions are shifted and nondimensionalized by subtracting a background ``climatological'' mean and dividing by the corresponding standard deviation on a per-variable and per-level basis\footnote{Unlike a standard climatology, the means and standard deviations here are taken over the whole period, without regard for time of day or day of year.}.  The model's output is similarly adjusted, being multiplied by the standard deviation of six-hour differences to form the forecast increment.  Likewise, the standard deviations of differences form the $\sigma^2_{\Delta\mathrm{var}}(k)$ factor in \eqref{eqn:trainerr}, allowing the aggregation of errors with different natural units and scales.

This nondimensionalization does not change the model's expressiveness, but it improves its numerical conditioning.  In nondimensional form, the model receives inputs and produces outputs that resemble draws from a unit normal distribution, without the need to learn large or small parameters to perform such adjustments internally.

However, these normalization constants depend on the distribution of the training set.  Although both ERA5 and GDPS model the same atmosphere, there are some systematic differences between the outputs of the two models.  Most critically, the GDPS and ERA5 datasets have considerable differences in their representation of stratospheric (above 100\hpa) specific humidity, with a different mean structure and considerably greater variability in the former.  

The baseline GraphCast model produces poor results when initialized with GDPS fields, most notably with large and accumulating errors in the stratosphere, motivating the fine-tuning process of this work.  Early fine-tuning experiments that did not adjust these normalization factors resulted in slow learning, and examination of the per-variable components of the \eqref{eqn:trainerr} loss showed that the model struggled to reduce the errors related to this difference in datasets.

The normalization factors used by \citet{graphcast} were computed over 1979--2015, a period sufficient to average out most interannual variability.  Unfortunately, while the GDPS dataset has systematic differences, the historical record is much shorter, and the corresponding normalization factors were computed over the July 2019--July 2021 period, cutting off the last few months of the training set to avoid introducing seasonal bias.

The scaled differences between the GDPS and ERA5 normalization factors are shown in figure \ref{fig:dstd_6h} alongside a control computation that computes the ERA5-based normalization factors over the same period.  Interannual variability is responsible for modest differences between the newly-computed and original normalization factors, but the stratospheric specific humidity remains a clear outlier in the GDPS data.

For the GDPS fine-tuning, these revised normalization factors were used in the first fine-tuning stage (``1a'', in table \ref{tab:schedule}), training over single-step forecasts for 2560 batches (10240 samples).  The normalization factors are normally in equilibrium with at least the input and output weights of the model, and replacing the normalization factors breaks this equilibrium.  Consequently, the method of section \ref{subsec:lr} could not be used to determine an optimal learning rate.  Instead, a peak learning rate of $1.25\cdot10^{-4}$ was used to encourage adaptation.

Since the control run utilized the ERA5 dataset, it did not need replacement normalization factors and the original ones were used.  However, single-step training effectively ``detunes'' the model from its prior multistep optimum, so the control run was still trained with the same learning rate schedule to characterize this effect.

\subsection{Replacing pressure-based error weights}\label{subsec:errweights}


During principal training of GraphCast, \citet{graphcast} used a simple functional form for the level-based error weight $w(k)$, of:
\begin{equation} \label{eqn:pressure_weight}
    w(k) = \frac{\mathrm{pressure}(k)}{\sum_{k=1} \mathrm{pressure}(k)}.
\end{equation}

Their documentation describes this as giving each level a weight approximately proportional to its density, with the intention of assigning greater importance to levels close to the surface.  However, this measure is unsatisfying for three reasons:
\begin{enumerate}
    \item This weighting fails to converge towards a proper (mass or height-weighted) vertical integral because it neglects layer thicknesses.  Model levels are not spaced uniformly in either height or pressure coordinates.  A hypothetical 38-level version of GraphCast that cloned an arbitrary model level would have no new information content or predictive power, but the pressure weighting of equation~\ref{eqn:pressure_weight} would double the influence of the cloned level in the loss function.  Compared to the 13-level version of the model, the 37-level version has finely-spaced levels in the lower troposphere, so \eqref{eqn:pressure_weight} might give undue weight to these levels at the expense of upper levels.
    \item This weighting assigns a disproportionately small fraction of the error to the stratospheric levels.  By construction, the 1\hpa level receives 0.0064\% of the integrated error weight in the 37-level version of GraphCast.  If the level is so unimportant that it receives so little weight, it would seem more sensible to simply cut it and have a slightly smaller, more computationally-efficient model.
    \item Finally, the ultimate goal of a forecasting system like GraphCast is to make accurate predictions over the medium-term, beyond the period covered directly by training.  Since the atmosphere is well-mixed, it's not obvious that the \emph{short-term} accuracy near the surface is the most accurate predictor of medium and longer-term performance.
\end{enumerate}

Additionally, changes to the $\sigma_{\Delta\mathrm{var}}^2$ factors implicitly change the weight given to the variable and level during training.  For example, compared to GraphCast's principal training, the revised increment standard deviations in figure~\ref{fig:dstd_6h} would tend to reduce the emphasis given to stratospheric specific humidity by a large factor (about $6$).

To address these factors, this work derives new level-based error weights through a sensitivity analysis, described in detail in appendix~A.  In brief, it measures the change in forecast output at 5-day lead times, on a per-level basis, when perturbing the initial condition towards that produced by a 6-hour forecast\footnote{This ensures that the perturbations are physically plausible, without requiring an external ensemble.}.  Taking a forecast initialized on 1 Jan 2020, 12:00 UTC as an example, the input perturbations would be derived by taking a 6-hour forecast initialized on 1 Jan 2020, 6:00 UTC, and the forecast change would be measured on 6 Jan 2020, 12:00 UTC.  The level sensitivity proportional to the ratio of output perturbation and input perturbations, separated by forecast level, and that sensitivity is taken to be the level weight in \eqref{eqn:trainerr} after normalization.

GraphCast's execution is nondeterministic at the level of floating-point round-off because of dynamic scheduling of its computational kernels, so there is some inherent randomness to this procedure such that a small enough sensitivity might be indistinguishable from random noise.  To measure this effect, an unperturbed run was added to the output set; the computed relative sensitivities for the perturbed run were all comfortably larger than could be caused by random variation.

\begin{figure}[tb]
    \begin{center}
        \includegraphics[width=6in]{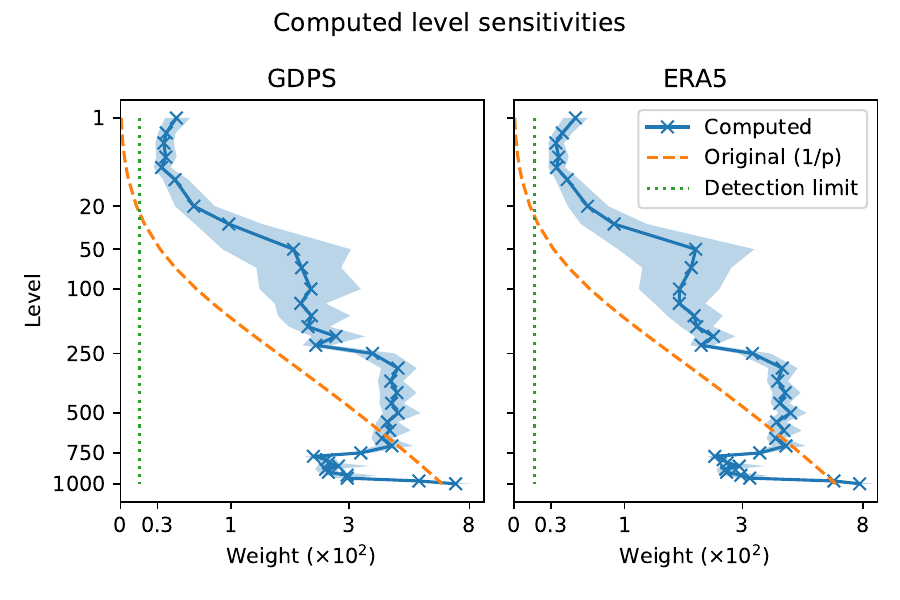}
    \end{center}
    \caption{Per-level perturbation sensitivities computed for the GDPS (left) and ERA5 (right), normalized so that the sum over all levels is $1$.  The shaded region shows bootstrapped confidence intervals at the 5th and 95th percentiles.}\label{fig:level_weights}
\end{figure}

Since the training datasets are different and the variable/level normalization factors are different, this procedure will result in different sensitivities and thus error weights for the GDPS and ERA5 training sets.  The computed weights are shown in figure~\ref{fig:level_weights}, and the respective weights were used for the GDPS fine-tuning and the ERA5 control training runs.

These weights show some decay towards higher levels and consequently lower pressure, but the overall functional form does not resemble the originally-prescribed $1/p$ scaling.  Instead, the calculated weighting is much more closely related to layer thickness, with the plateaus in figure \ref{fig:level_weights} corresponding to compressed spacing of levels in the lower troposphere, relatively increased spacing in the middle troposphere, and (necessarily) reduced spacing again in the upper troposphere and stratosphere.  Weighting based on layer thickness would correspond to a straightforward vertical integral, but the lowest and highest levels still show sensitivities several times greater than would be predicted by this approximation.

With these vertical error weights determined, a second training stage over one-step forecasts was undertaken.  Heuristically speaking, one-step forecast performance will be at its maximum after this stage, with subsequent multistep training acting more as a compromise between single-step and longer-term forecast accuracy.  In addition, training over one-step forecasts is relatively quick compared to multistep forecasts, so this stage consisted of $20480$ batches ($81920$ training samples), and the method of section~\ref{subsec:lr} was used to give a peak learning rate of $3.75\cdot10^{-7}$. 

\subsection{Training schedule}\label{subsec:schedule}

Principal training of GraphCast extended the forecast from one step (6\hr) to twelve steps (72\hr) one step at a time, with a fixed learning rate ($3 \cdot 10^{-7}$) and 1000 training batches (32000 samples).  This is comprehensive, but it seems overly cautious.  Larger jumps in forecast length are supportable provided GraphCast trained on $N$ steps already provides a reasonably good forecast at further lengths, and this is indeed the case during the fine-tuning procedure.  Some intermediate training steps can be skipped, redirecting the total computational time towards longer forecasts.

This work limits the training to 1, 2, 4, 8, and 12-step forecasts, increasing the forecast length first to one day and then incrementing the length by one day at a time.  

The overall learning rates, training lengths, and training time and shown in table~\ref{tab:schedule}.  The final choice of 81920 training samples for 1-step forecasts (with the revised error weights), 20480 training samples for 2-step forecasts, and 10240 samples for 4-step forecasts approximately equalizes the training time between these stages.  For all stages of training, the initial learning rates of section~\ref{subsec:lr} were significantly larger than the learning rate used for GraphCast's principal training, adjusted for the smaller batch size.

\subsection{Split-horizon training}\label{subsec:split}

The final training stage, over 12-step (3-day) forecasts required adjustment.  Although the system used could successfully backpropagate over the full forecast window, there was not enough host memory to perform the gradient calculations while loading the next batch's data in parallel.  Without this parallelism, training would have taken significantly longer and been an inefficient use of the system's GPUs.

This problem is shared with some other forecast models.  Pangu-Weather \citep{pangu} trained separate models to predict forecast increments at 1, 3, 6, and 24-hour lead times, avoiding autoregressive training.  The FengWu model \citep{fengwu} extends its one-step training to longer lead times through a replay buffer, where a forecast after $N$ steps (varying) is written to disk and later used as initial conditions for the $N+1$-step forecast.  This essentially breaks up the autoregressive training into one-step pieces, where gradients do not flow between steps.

\begin{figure}[tb]
    \begin{center}
        \includegraphics[width=6in]{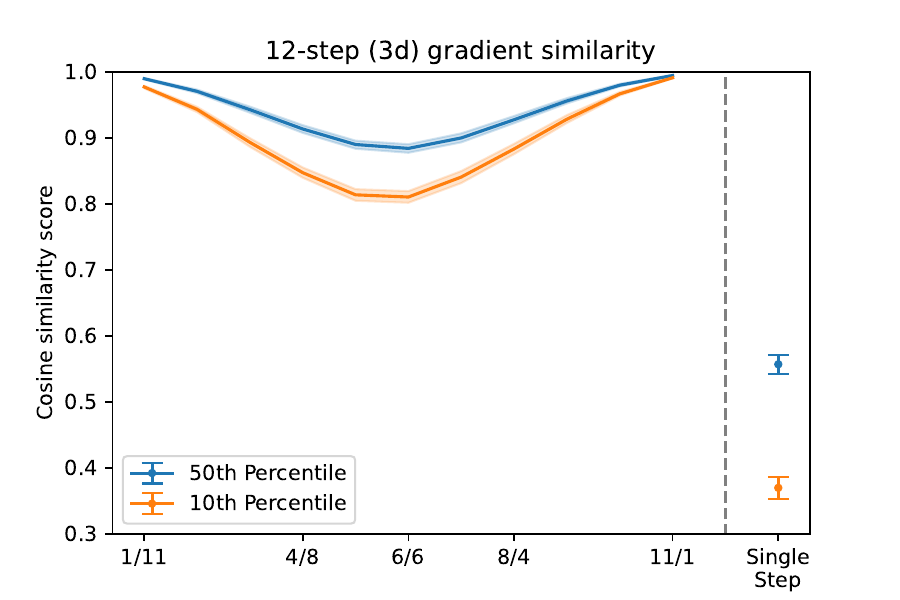}
    \end{center}
    \caption{Mean cosine similarity (componentwise correlation) at tenth and fiftieth percentiles between gradients computed over 12 steps (72\hr) and those computed by splitting the computation into several pieces.  Curves: $1/11$ through $11/1$ split; points: $1/1/\cdots/1$ split (individual steps), like \citet{fengwu}.  The error bars and shaded regions mark the 5th and 95th percentiles via bootstrapping. The computation was performed over 40 random forecast dates from the training set, using the ERA5 database and the baseline 37-level GraphCast model.}\label{fig:horizon}
\end{figure}

This work takes inspiration from the replay buffer.  Rather than freeze initial conditions to perform training over a single step, however, the 12-step forecast is split into just two stages of 4 and 8 steps respectively, with the latter calculation initialized from the output of the 4-step forecast\footnote{The initial conditions were also computed in 32-bit precision, more closely corresponding to the model's intended use in production.}.  The gradients and loss are accumulated over both stages, so the ultimate loss function is unchanged from equation~\ref{eqn:trainerr}.  This reduces the host memory requirement, both through limiting the memory required for the gradient computation and by separating the data-loading task into its subcomponents.  This separation is logically equivalent to using the ``detach gradient'' operator in PyTorch.

Splitting the training horizon in this way would ideally result in aggregated gradients indistinguishable from gradients computed over a full 12-step forecast.  However, this is not guaranteed.  Figure~\ref{fig:horizon} shows the impact of splitting 12-step training into two steps, varying the split point, by evaluating the cosine similarity of gradients by parameter set and showing the lowest correlation figure among the model's parameter sets. 

The cosine similarity between two vectors is their componentwise correlation or length-normalized dot product, and the name ``cosine similarity'' comes from the observation that the angle between two vectors is the cosine of this quantity.  This was chosen as the evaluation metric because momentum-based optimizers like AdamW can adjust for a re-scaled gradient, computing the same updates despite the scaling.  As long as the correlations remain high, stochastic gradient descent is likely to follow the same path of convergence in parameter space.  However, if these correlations fall, it's not clear that the split-horizon model would converge towards the same solution as the unitary model.

Overall, the gradients computed by splitting the model training into two steps remain similar to those computed over the full forecast length, and the correlation improves for divisions that maximize the length of the longer steps.  However, accumulating the gradients over single steps alone, as in the FengWu replay buffer, shows significantly reduced correlation.

\section{Results}\label{sec:results}

\subsection{Validation loss}

\begin{figure}[tb]
    \begin{center}
        \includegraphics[width=6in]{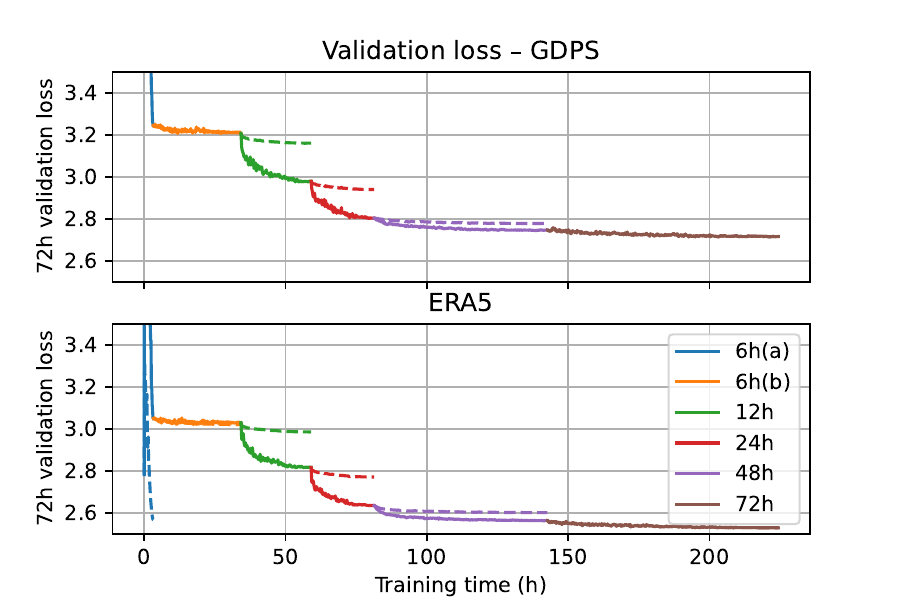}
    \end{center}
    \caption{Validation loss over 72\hr for all stages of training, for GDPS fine-tuning (top) and the control ERA5 fine-tuning (bottom), versus cumulative training time.  The dashed lines indicate each stage's native validation loss (e.g. 2-step validation loss for 12\hr forecasts), scaled to begin at the same point as stage's 72\hr validation loss.}\label{fig:vloss}
\end{figure}

The simplest error measure is the validation loss over training, using the loss function~\eqref{eqn:trainerr}.  This validation loss was computed over 64 dates randomly selected from the validation period.  Figure~\ref{fig:vloss} shows the 72\hr (12-step) loss evaluated over the full fine-tuning runs, showing the progression of overall error.  

Since only the last stage was trained with this loss function, the figure also shows with dashed lines the evolution of each stage's own (N-step) validation loss, scaled to the 72\hr loss value.  These lines are most visible for the 12\hr, 24\hr, and 48\hr forecast stages.  The difference between the dashed and solid lines corresponds to the additional gain over 72\hr compared to the trained period. For example, the 12\hr training stage causes an 0.5\% reduction in the validation loss when measured over 12 hours, but it causes a much larger 5\% reduction in the validation loss when measured over 72\hr.  This suggests that relatively subtle differences in the short-term forecast have larger impacts over the medium and longer terms, although it's not immediately clear whether the resulting improvements come from a better representation of systematic trends or undesirable smoothing of shorter scales.

The GDPS and control runs show similar behaviour for multistep forecasts, but figure~\ref{fig:vloss} shows an interesting difference in the very first phase of training.  There, the one-step validation loss for the GDPS fine-tuning is essentially proportional to the multistep validation loss, but the one-step validation loss for the ERA5 control run eventually decreases below its starting value while the multistep loss remains elevated.  This is the ``de-tuning'' expected from taking model weights initially trained over several steps and training them over a single step: the model forgets a portion of the systematic corrections that help over longer forecasts.  

Since the GDPS fine-tuning does not show this behaviour, it suggests that the shock of adapting to the systematic differences is larger than the portability of any multistep forecasting skill retained from the initial model weights.

\subsection{Error versus analysis}

\begin{figure}[tb]
    \begin{center}
        \includegraphics[width=\textwidth]{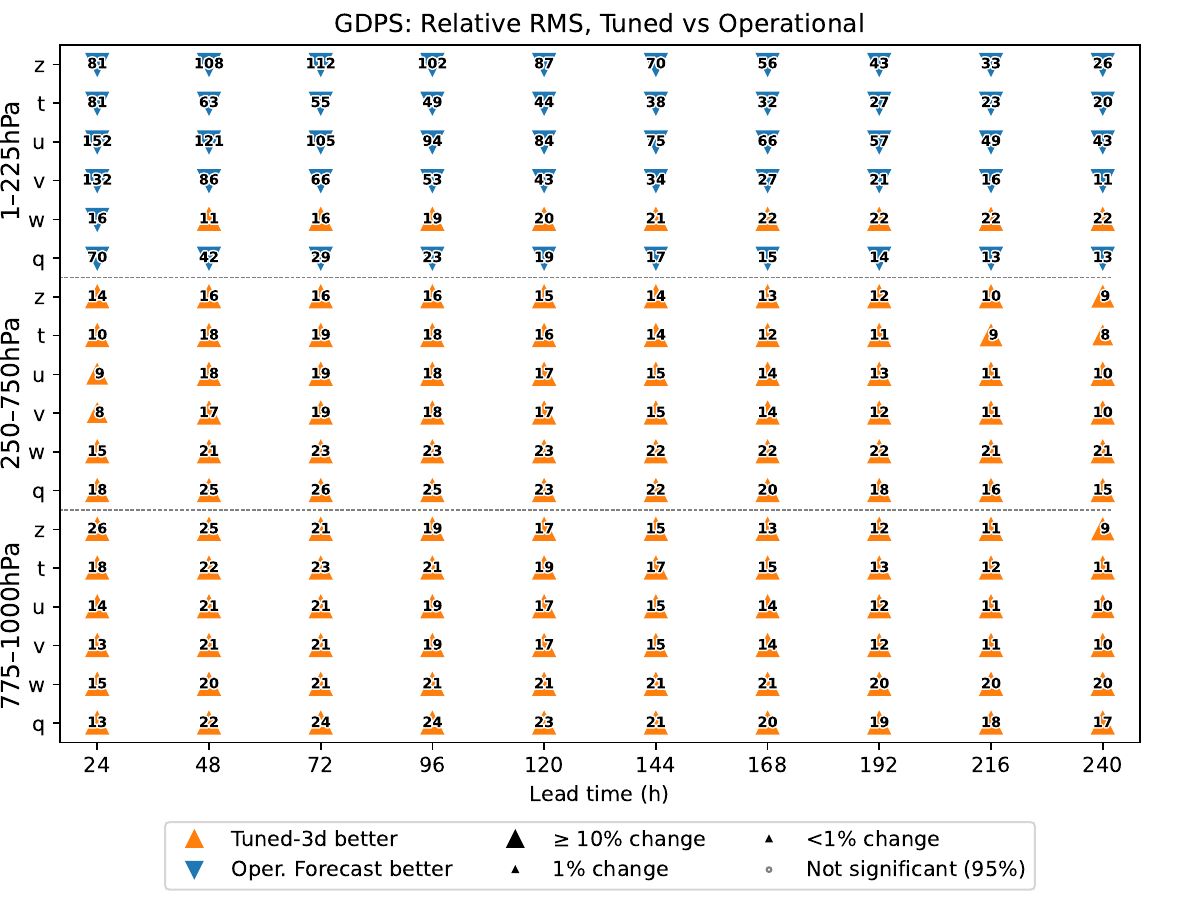}
    \end{center}
    \caption{``Scorecard'' of relative RMS error of the fine-tuned GraphCast versus the operational deterministic forecast for calendar year 2023, using forecasts initialized at 0Z and 12Z with skill scores averaged over the indicated level ranges.  Upwards pointing triangles indicate that the fine-tuned GraphCast is better (lower error) than the operational forecast over its indicated variable, level range, and lead time; downwards pointing triangles indicate worse performance.  Overlain numbers show the relative percentage of improvement or degradation when the magnitude is larger than 5\%.}\label{fig:scorecard_g1}
\end{figure}

\begin{figure}[tb]
    \begin{center}
        \includegraphics[width=6in]{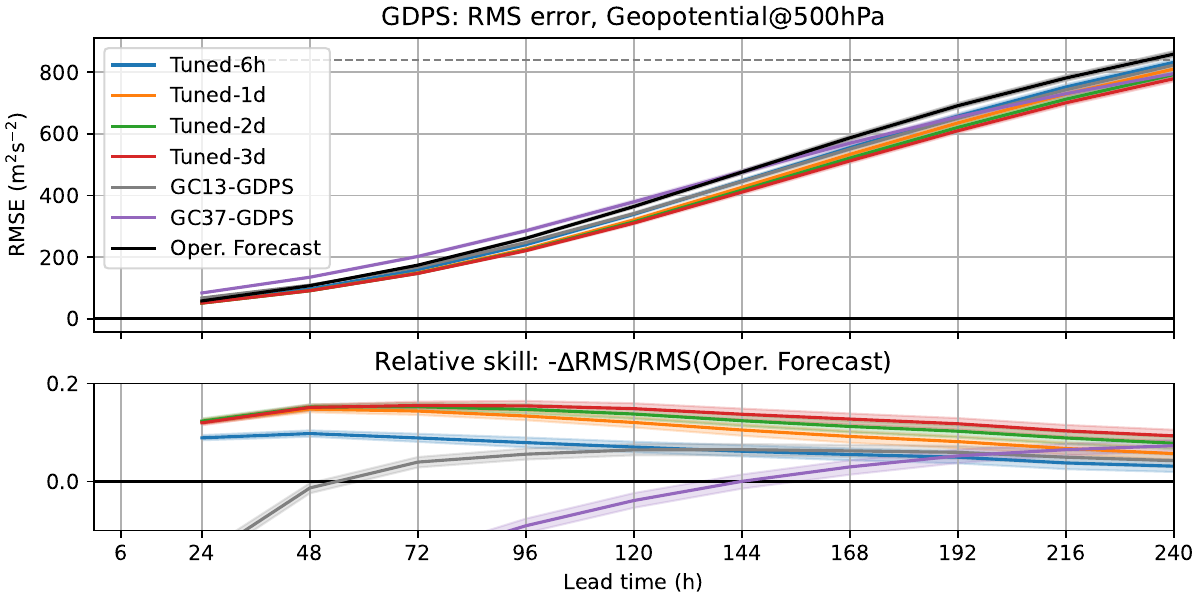}
    \end{center}
    \caption{Evolution of the RMS error against the GDPS analysis over lead time for geopotential at the 500\hpa level, for unmodified GraphCast-37 and GraphCast-13, each stage of fine-tuning, and the operational forecast.  The thin dashed line is the climatological error level.  The lower panel shows the skill score relative to the operational forecast, emphasizing the differences between the data-driven models.}\label{fig:z500_rms_g1}
\end{figure}

\begin{figure}[tb]
    \begin{center}
        \includegraphics[width=\textwidth]{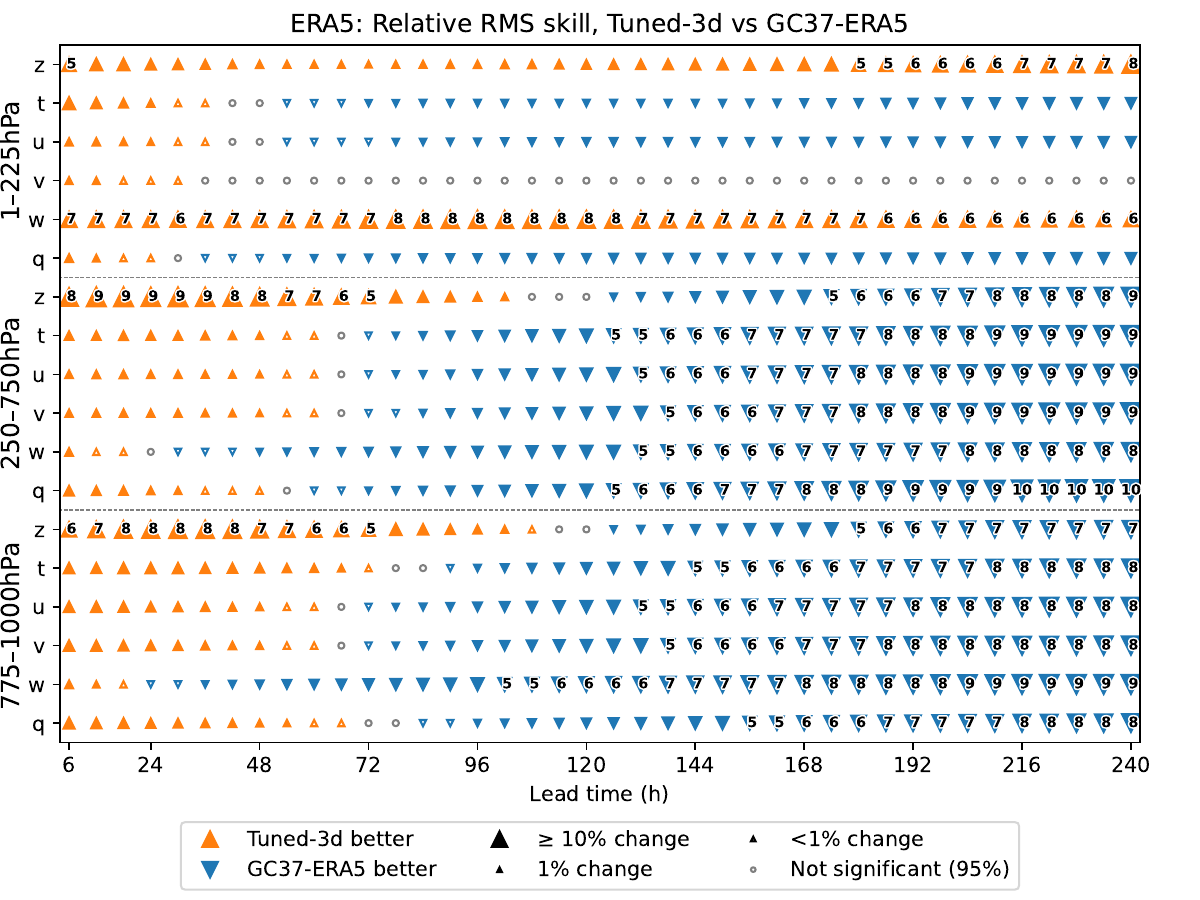}
    \end{center}
    \caption{As figure~\ref{fig:scorecard_g1}, for the ERA5 fine-tuned run versus the unmodified GraphCast-37 weights over calendar year 2023 forecasts initialized every six hours.}\label{fig:scorecard_era5}
\end{figure}

A closer look at the error by meteorological variable, lead time, and level shows additional detail in the residual errors of fine-tuned GraphCast.  

Figure \ref{fig:scorecard_g1} shows the root mean squared (RMS) errors of the GDPS-tuned GraphCast at the end of the training process in section \ref{sec:method} against the corresponding error of ECCC's operational forecast over calendar year 2023, interpolated from a 15\unit{\km} notional resolution to $0.25$°.  The operational forecast was initialized twice daily, at 0Z and 12Z, using initial conditions based on the operational analysis, which has a +3\hr data cutoff.  This operational dataset and the training dataset have relatively small differences.

The fine-tuned GDPS improves over the operational forecast at every lead time in the troposphere (below 250\hpa) and planetary boundary layer, and this improvement is most significant for the final weights produced by the fine-tuning process.  Figure \ref{fig:z500_rms_g1} shows the RMS error of geopotential at 500\hpa for each set of the tuned GraphCast weights.  For this variable, the overall error of the previous-best GraphCast version, the 13-level HRES-tuned version, was matched after the first (6\hr) tuning step.

For the control run with ERA5 data, the results are much more mixed.  The evaluation of the final fine-tuned model against unmodified GraphCast-37 is shown in figure \ref{fig:scorecard_era5}, for forecasts initialized every six hours and valid in calendar year 2023.  The fine-tuned version shows modest improvements over most variables and levels up to its 72\hr training window, but performance is degraded thereafter.  The 48\hr version (not shown) has better performance when evaluated over 2023, 

\begin{figure}[tb]
    \begin{center}
        \includegraphics[width=6in]{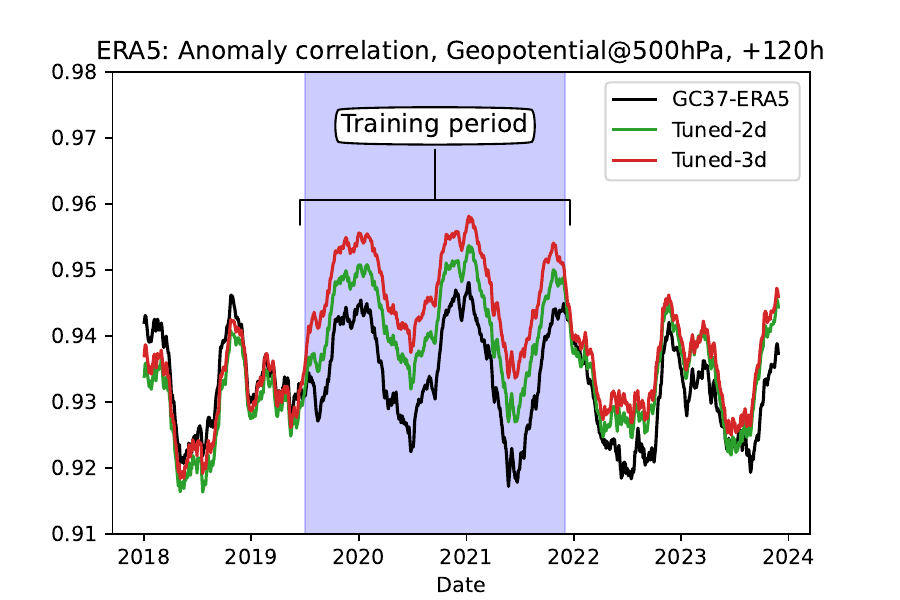}
    \end{center}
    \caption{Anomaly correlation against the ERA5 reanalysis of geopotential at 500\hpa at +120\hr (5\dd) lead time over forecast initialization date, for the unmodified 37-level GraphCast and fine-tuned model after the 2 and 3-day stages.  The training period is shaded.  For clarity, the anomaly correlations are smoothed over a 30-day sliding window.}\label{fig:overfit}
\end{figure}

To evaluate the over-fitting hypothesis, the anomaly correlation\footnote{The anomaly fields subtract the ERA5 climatological mean from 1990 to 2019 by day of year and time of day, computed in \citet{weatherbench2}.} of geopotential at 500\hpa and 120\hr lead time is shown in figure \ref{fig:overfit} from 2017 – before the training period for fine-tuning but after the primary training period for GraphCast – and the end of 2023 for the fine-tuned versions (48\hr and 72\hr) from the control run and the unmodified 37-level GraphCast, all initialized from the ERA5 dataset.  The anomaly correlation for both fine-tuned versions is elevated over the training period, but only a portion of this improvement persists afterwards and into the 2022 validation and 2023 testing years.

Interestingly, the fine-tuned model versions do not show improvement relative to the unmodified versions for times before the training period, although it is equally out-of-sample as the validation and test periods.  This may provide weak evidence for the hypothesis that more recent analyses are better-constrained by observational data and expose more predictability that data-driven models can learn from, since information obtained by observations first assimilated in 2020 or 2021 would not be present in the reanalysis fields of 2018 or 2019.

\begin{figure}[tb]
    \begin{center}
        \includegraphics[width=6in]{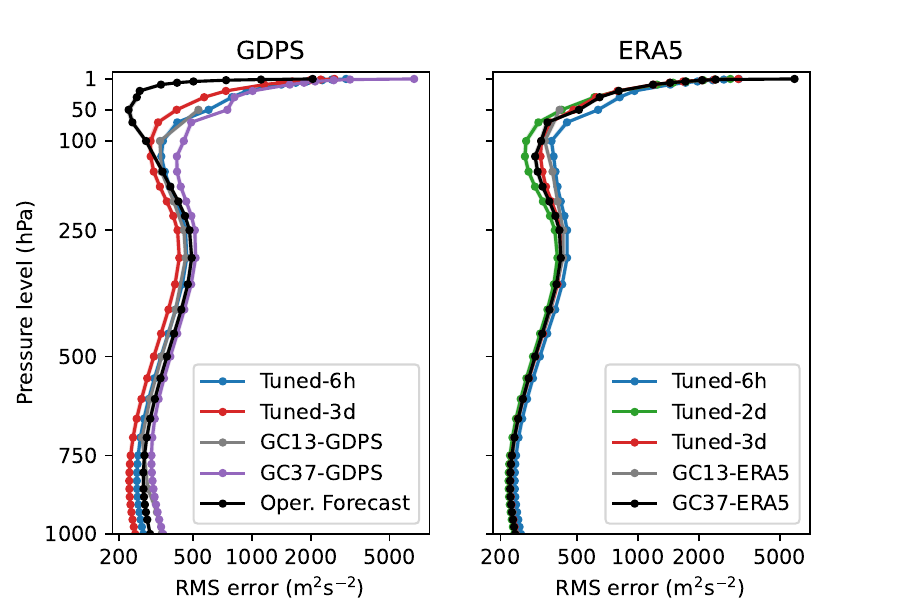}
    \end{center}
    \caption{RMS error for geopotential at +120\hr (5\dd) by vertical level, over the GDPS fine-tuning process (left) and control ERA5 fine-tuning process (right), evaluated against the respective analyses.  When fine-tuning against the GPDS dataset, the vertical profile of GraphCast's error remains similar to the originally-trained version, and the overall error is better than that of the operational forecast below 100\hpa.} \label{fig:z_vertical_rms}
\end{figure}

For both versions of the fine-tuned model, figure \ref{fig:z_vertical_rms} shows the RMS error in geopotential by vertical level at +120\hr lead time.  

Compared to the unmodified GraphCast model, the fine-tuned model significantly reduces the geopotential error at all vertical levels when the model is initialized with the GDPS analysis.  This error reduction is slightly larger in the stratosphere than at lower levels, suggesting that the fine-tuned model has largely adapted to the dramatically different stratospheric humidity fields in the GDPS analysis.  Additionally, comparison against the operational forecast (left panel) shows that the vertical structure of GraphCast's error remains consistent with itself over the fine-tuning process while the magnitudes decrease, suggesting that the fine-tuned models continue to ``remember'' training skill gained from GraphCast's primary training.

The model fine-tuned on ERA5 data generally reproduces the error magnitudes and structure of the unmodified model, although there is a noticeable reduction in RMS error at the topmost (1\hpa) level.  This might be the result of the modified level weighting of section~\ref{subsec:errweights}.

\begin{sidewaystable}
    \begin{center}
        \begin{tabular}{cc|ccc|ccc|ccc|ccc}
            & Model & \multicolumn{3}{c}{z 500\hpa (\unit{\m^2\per\s^2})} 
                  & \multicolumn{3}{c}{t 850\hpa (\unit{\K})} 
                  & \multicolumn{3}{c}{q 700\hpa (\unit{\g\per\kg})} 
                  & \multicolumn{3}{c}{(u,v) 850\hpa (\unit{\m\per\s})} \\
            & & 1\dd & 5\dd & 10\dd & 1\dd & 5\dd & 10\dd & 1\dd & 5\dd & 10\dd & 1\dd & 5\dd & 10\dd \\  \hline
            \multirow{7}{*}{\rotatebox{90}{ERA5}} &
              GC13-ERA5  & 46.99 & 305.54 & 789.25 & 0.72 & 1.79 & 3.59 & 0.65 & 1.27 & 1.80 & 1.88 & 5.11 & 8.94  \\
            & GC37-ERA5  & 47.69 & 305.74 & \textbf{745.49} & 0.64 & 1.77 & \textbf{3.43} & 0.62 & \textbf{1.21} & 1.69 & 1.79 & \textbf{4.94} & 8.39  \\
            & Tuned-6h  & 45.32 & 323.59 & 825.61 & 0.63 & 1.94 & 3.83 & 0.62 & 1.37 & 1.94 & 1.78 & 5.48 & 9.40  \\
            & Tuned-1d  & 43.23 & 303.99 & 797.22 & \textbf{0.61} & 1.81 & 3.67 & 0.60 & 1.29 & 1.89 & 1.72 & 5.21 & 9.16  \\
            & Tuned-2d  & \textbf{43.20} & \textbf{298.11} & 783.79 & 0.61 & \textbf{1.76} & 3.61 & \textbf{0.60} & 1.23 & 1.82 & \textbf{1.72} & 5.08 & 9.00  \\
            & Tuned-3d  & 43.52 & 307.27 & 809.93 & 0.62 & 1.83 & 3.72 & 0.61 & 1.27 & 1.85 & 1.74 & 5.17 & 9.12  \\
            & Climatology  & 836.15 & 836.15 & 836.15 & 3.54 & 3.54 & 3.54 & 1.68 & 1.68 & \textbf{1.68} & 8.08 & 8.08 & \textbf{8.08}  \\
            \hline
            \multirow{8}{*}{\rotatebox{90}{GDPS}} &
              GC13-GDPS  & 67.16 & 341.25 & 823.21 & 1.28 & 2.30 & 3.88 & 0.78 & 1.33 & 1.86 & 2.40 & 5.49 & 9.21  \\
            & GC37-GDPS  & 83.54 & 379.44 & 795.80 & 1.35 & 2.38 & 3.74 & 0.79 & 1.39 & 1.83 & 2.52 & 5.77 & 8.83  \\
            & Oper.~Forecast  & 57.81 & 364.25 & 859.09 & 0.84 & 2.15 & 3.91 & 0.81 & 1.63 & 2.12 & 2.44 & 6.28 & 9.92  \\
            & Tuned-6h  & 52.67 & 338.96 & 832.42 & 0.75 & 1.97 & 3.82 & 0.73 & 1.52 & 2.15 & 2.22 & 5.80 & 9.59  \\
            & Tuned-1d  & 50.76 & 320.77 & 810.68 & 0.70 & 1.83 & 3.66 & 0.69 & 1.37 & 1.97 & 2.13 & 5.42 & 9.27  \\
            & Tuned-2d  & \textbf{50.68} & 314.44 & 792.58 & \textbf{0.70} & 1.79 & 3.58 & \textbf{0.69} & 1.31 & 1.88 & \textbf{2.12} & 5.28 & 9.05  \\
            & Tuned-3d  & 50.88 & \textbf{310.48} & \textbf{779.26} & 0.70 & \textbf{1.75} & \textbf{3.50} & 0.69 & \textbf{1.27} & 1.81 & 2.13 & \textbf{5.16} & 8.86  \\
            & Climatology  & 839.39 & 839.39 & 839.39 & 3.65 & 3.65 & 3.65 & 1.72 & 1.72 & \textbf{1.72} & 8.10 & 8.10 & \textbf{8.10}  \\
        \end{tabular}
    \end{center}
    \caption{RMS losses for a selection of variables and lead times, by model and ground-truth dataset.  The ``Tuned-'' rows correspond to the model versions trained in this work.  The best results are highlighted in bold.}\label{tab:weatherbench}
\end{sidewaystable}

Table \ref{tab:weatherbench} presents quantitative RMS errors for a selection of variables and lead times, a subset of the ``headline'' statistics from \citet{weatherbench2}, with the best error per lead time and variable in bold.  

\subsection{Smoothing}

\begin{figure}[tb]
    \begin{center}
        \includegraphics[width=6in]{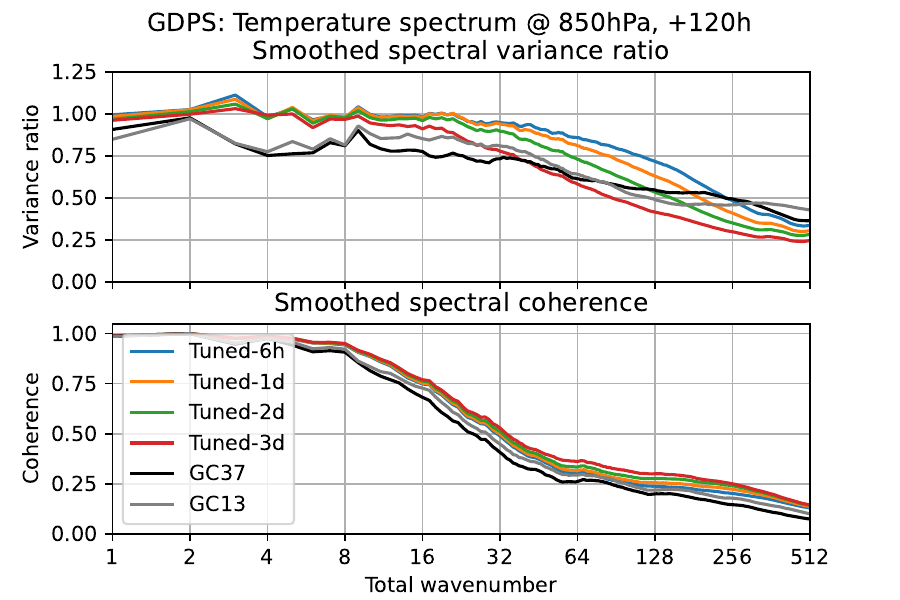}
    \end{center}
    \caption{Spectral variance ratio (top) and coherence (bottom) versus the GDPS analysis for temperature at 850\hpa and +120\hr (+5\dd) lead time, for the model at each stage of fine-tuning and for the unmodified 13-level and 37-level versions.  The shown lines are mean values over forecasts initialized from 50 random dates in the test period.  For visual clarity, the values are averaged over wavenumbers within 10\% of the indicated value; e.g. the value at wavenumber 32 includes wavenumbers 29 through 35.} \label{fig:spectrum_gdps}
\end{figure}

\begin{figure}[tb]
    \begin{center}
        \includegraphics[width=6in]{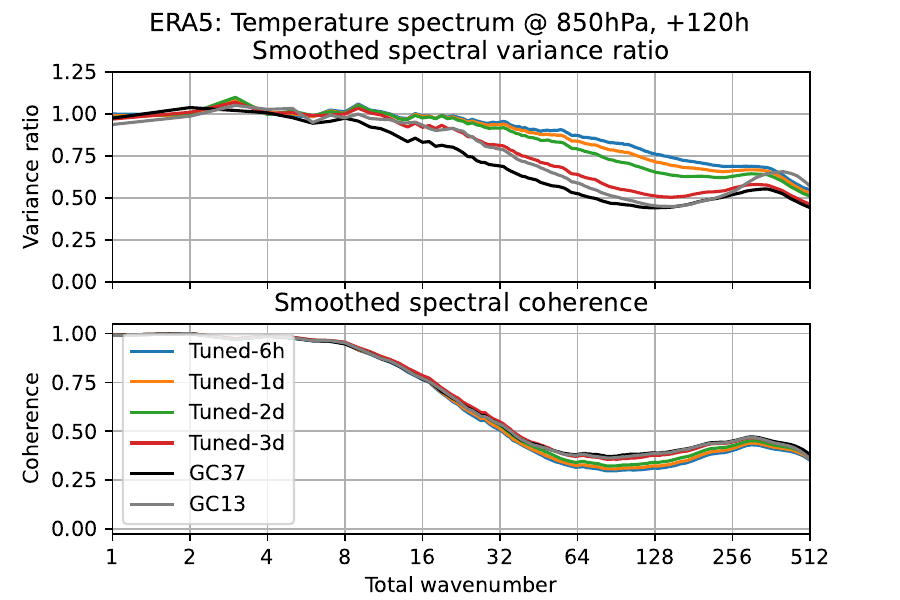}
    \end{center}
    \caption{As figure~\ref{fig:spectrum_gdps}, for the ERA5 control runs.} \label{fig:spectrum_era5}
\end{figure}

Deterministic data-driven forecast models trained to minimize mean squared errors calculated like equation \eqref{eqn:trainerr} are well-known to suffer from smoothing of fine scales (high wavenumbers) at longer lead times.  The intuition here is that finer scales are less predictable than synoptic scales at long lead times, and so the model learns to suppress these fine scales rather than suffer a ``double-penalty'' from realistic variation that is misaligned compared to the ground-truth analysis.

The various training stages of section \ref{sec:method} allow visualization of this phenomenon as training progresses.  Figures \ref{fig:spectrum_gdps} and \ref{fig:spectrum_era5} show the evolution of the spectral variance ratio (variability by total spherical harmonic wavenumber relative to the ground-truth analysis) and coherence (correlation by wavenumber) for the GDPS fine-tuning and control cases, respectively.  In each case, smoothing of fine scales increases as the training continues, although the coherence also increases.  

The increase in smoothing is particularly notable when comparing the 2\dd-trained and 3\dd-trained versions.  On the other hand, the 6\hr-trained model remains relatively sharp, suggesting that alternative error functions like those used by \citet{neuralgcm} are most needed when training over long forecast lengths.

Comparing figures \ref{fig:spectrum_gdps} and \ref{fig:spectrum_era5}, the GDPS-tuned model shows more smoothing at wavenumbers above 50, with the variance ratio falling to 0.25 at wavenumber 512 compared to 0.5 for the ERA5-tuned model.  This might be a consequence of the comparatively lower coherence at these scales, since removing unpredictable scales tends to improve the mean squared error.

\section{Discussion \& Conclusion}\label{sec:conclusion}

This work demonstrates that effective fine-tuning of GraphCast is possible, adapting the model to a different analysis system while using a small fraction of the computational time and training data used to train GraphCast from scratch.  Adopting split-horizon training allowed effective computation of gradients over 3\dd forecasts (12 6\hr steps) despite memory limitations on the computer system used for the fine-tuning.  Although the training period used here was a sample of convenience based on updates to ECCC's GDPS, the relatively short training period was nonetheless enough to produce a model that outperformed the operational forecast.

The resulting fine-tuned model improves upon ECCC's operational forecast at $0.25$° resolution, although like the unmodified GraphCast model it smooths fine-scale structures at longer lead times.  This smoothing is particularly notable for the models trained on 3\dd forecasts, suggesting that future efforts to train models like GraphCast over longer forecasts may need to be particularly careful.  Training could follow in the steps of \citet{neuralgcm} to use alternative loss calculations that encourage sharpness.  

Alternatively, future work might adopt measures like low-rank adaptation \citep{lora} to reduce the number of degrees of freedom trained over longer forecasts.  This method trains low-rank updates to the model's weight matrices rather than allow their entries to vary freely over the gradient descent, and in language modelling the technique is used to fine-tune models towards new tasks while retaining baseline skill.  If this smoothing process is caused by the model over-correcting towards longer forecasts, then this technique might help preserve the relative sharpness of the model before long-forecast training.

A control model trained against the ERA5 reanalysis for the same period showed moderate levels of over-fitting, with the model having degraded performance over longer lead times in the out-of sample test period.  In-sample performance was also significantly better than out-of sample performance.  It's also possible that this smoothing is a necessary condition for GraphCast to over-fit the training set.  GraphCast has relatively few parameters relative to the raw number of degrees of freedom in a $0.25$° atmospheric analysis, but smoothing reduces the effective number of degrees of freedom.  Training that preserves sharpness may also help reduce over-fitting, improving generalization from limited training data.

Future work will focus on these aspects of model training, both to determine how much data is required to create an effective data-driven forecast model and to explore ways of preserving sharp forecasts over long-forecast training.

\section*{Code and data availability}

The ERA5 data used for fine-tuning is freely available from the WeatherBench~2 dataset, with instructions for access at \url{https://weatherbench2.readthedocs.io/en/latest/data-guide.html}.  The GDPS forecast and analysis data is very large (about 6.5 TB), but access can be arranged upon request.

The Python source code used for model training is available at \url{https://github.com/csubich/graphcast}, under the Apache license.  The model checkpoints produced from the fine-tuning processes are available at \url{https://huggingface.co/csubich/graphcast_finetune_2019_2021}.  Since these checkpoints are based on the original GraphCast model weights of \citet{graphcast}, they  retain the Creative Commons Attribution-NonCommercial-ShareAlike 4.0 license (\url{https://creativecommons.org/licenses/by-nc-sa/4.0/}).

\bibliographystyle{ametsocV6}
\bibliography{paper_ml}

\clearpage

\appendix

\begin{algorithm}[t]
    \begin{algorithmic}
        \State Select 64 dates $D_i, i \in [1,\ldots,64]$ from the training period
        \For {$d \in D$}
            \State Load initial conditions $x(i,j,k ; d - 12\hr)$, $x(i,j,k ; d - 6\hr)$, and $x(i,j,k;d)$
            \State Compute control forecast $\hat{p}(i,j,k,t \in (d+6\hr \ldots d+5\dd); x(d-6\hr), x(d))$ 
            \Statex initialized at date $d$ and valid for 5 days
            \State Compute perturbation forecast $p'(i,j,k,t = d; x(d-12\hr), x(d-6\hr))$
            \For {$\tilde{k} \in [0,37]$}
                \If{$\tilde{k} \neq 0$} \Comment{Compute perturbed initial conditions}
                    \LComment{Measure the level-unweighted MSE of each level $k$ in the perturbation reforecast}
                    \State $ERR_{\tilde{k}} \gets$ \Call{MSE}{$x(i,j,\tilde{k};d)$, $p'(i,j,\tilde{k},d)$}, where $w(\tilde{k})=1$ in \eqref{eqn:trainerr}
                    \LComment{Compute the perturbation weight necessary to have a computed MSE of 1}
                    \State $\epsilon \gets 1/ERR_{\tilde{k}}$
                    \State $x'(i,j,k;d) \gets (1-\epsilon \delta(k=\tilde{k}))x + (\epsilon \delta(k=\tilde{k})) p'(i,j,k,d)$
                \Else \Comment{Control to measure inherent randomness}
                    \State $x' \gets x$
                \EndIf
                \State Compute trial forecast $p'(i,j,k,t \in (d+6\hr \ldots d+5\dd); x(d-6\hr), x'(d))$
                \For {$day \in [1\dd,\ldots,5\dd], var \in (\text{msl, t2m, tp, z, t, q}), \kappa \in \{\text{GraphCast-13 levels}\}$}
                    \LComment{Output non-normalized mean square errors for a selection of variables and levels, at 1\dd to 5\dd lead times.}
                    \State \Output $\Vert p_{var}(i,j,\kappa,d+day) - p'_{var}(i,j,\kappa,d+day)\Vert^2$
                \EndFor
            \EndFor
        \EndFor
    \end{algorithmic}
    \caption{Algorithm to compute per-level sensitivity of a GraphCast forecast.}\label{alg:sensitivity}
\end{algorithm}

\section{Sensitivity analysis for error weights} \label{app:sensitivity}

The procedure of algorithm~\ref{alg:sensitivity} creates 210 sensitivity output points per forecast date and perturbed model level.  To compensate for different scales between variables and level, each of these output points is converted to its z-score over the initializing perturbations (both forecast date and level) by subtracting the mean and dividing by the standard deviation.  Taking the mean of these z-scores over the (output point, initializing date) dimensions and dividing by the global mean then gives the relative sensitivity of the output to each input level. 

\section{Statistical definitions}\label{app:stats}

For section \ref{sec:results}, the following definitions were used, generally following chapter 12 of \citet{ecmwf_user}:

\begin{itemize}
\item Given a scalar prediction $x_p(i,j,k)$ and corresponding analysis field $x_a$, the root mean squared error is simply:
\begin{equation} \label{eqn:rmse}
        RMSE(x_p,x_a; k) = \sqrt{\sum_i \sum_j \frac{\mathrm{d}A}{4\pi} (x_p(i,j,k) - x_a(i,k,k))^2}
\end{equation}
\item Given two prediction fields $x_p$ and $x_p'$ and one analysis field, the relative skill score of $x_p'$ versus $x_p$ is:
\begin{equation} \label{eqn:skill}
        SKILL(x_p', x_p, x_a ; k) = 1-\frac{RMSE(x_p',x_a;k)}{RMSE(x_p,x_a;k)}
\end{equation}
\item When a climatological field $x_c(i,j,k)$ is available, 
the activity is the root mean squared difference between a forecast or analysis field and the climatology:
\begin{equation} \label{eqn:act}
        ACT(x,x_c ; k) = \sqrt{\sum_i \sum_j \frac{\mathrm{d}A}{4\pi} (x(i,j,k) - x_c(i,k,k))^2}
\end{equation}

The activity of the analysis field is also the climatological error.
\item For a prediction and analysis field, the anomaly correlation is the correlation of the anomalies (field less climatological value).  Notably, this value does not have a bias removed:
\begin{equation} \label{eqn:acc}
        ACC(x_p, x_a, x_c ; k) = \frac{\sqrt{\sum_i \sum_j \frac{\mathrm{d}A}{4\pi} (x_p(i,j,k) - x_c(i,j,k)) (x_a(i,j,k) - x_c(i,j,k))}}
                                      {\sqrt{ACT(x_a,x_c;k) ACT(x_p,x_c;k)}}
\end{equation}
\end{itemize}

For the spectral decomposition shown in figures \ref{fig:spectrum_gdps} and \ref{fig:spectrum_era5}, the spectral decomposition of the forecast and analysis fields were taken with the S2FFT library of \citet{s2fft}.  To compute spherical harmonics up to a total wavenumber of 1079, the fields were first interpolated onto a uniform grid of size $2160 \times 2159$ using Fourier cosine interpolation in longitude and periodic Fourier interpolation in latitude.

S2FFT performs a complex-valued spherical harmonic transform, the cross spectral density is defined as:
\begin{equation}\label{eqn:specross}
    CROSS(x,y; \kappa) = \sum_\lambda \mathscr{S}(x)(\lambda, \kappa) \overline{\mathscr{S}(y)(\lambda, \kappa)},
\end{equation}
where $\mathscr{S}(x)(\lambda, \kappa)$ is the spherical harmonic transform with zonal wavenumber $\lambda$ and total wavenumber $\kappa$, and $\overline{(\cdot)}$ is the complex conjugate.  The spectral variance of a field is:
\begin{equation}\label{eqn:specvar}
    SVAR(x ; \kappa) = \vert CROSS(x,x ; \kappa) \vert,
\end{equation}
and the spectral coherence is:
\begin{equation}\label{eqn:speccoh}
    SCOH(x, y ; \kappa) = \frac{\vert CROSS(x,y) \vert^2}{SVAR(x) SVAR(y)}.
\end{equation}
Notably, the spectral coherence defined here is a ratio of variances, while the anomaly correlation coefficient of \eqref{eqn:acc} is a ratio of standard deviations.

\end{document}